\documentclass[conference]{IEEEtran}
\IEEEoverridecommandlockouts
\usepackage[
    pagebackref,
    breaklinks,
    hidelinks
]{hyperref}
\usepackage{epsfig}
\usepackage{bm}
\usepackage{multirow}
\usepackage{color}
\usepackage{booktabs} 
\usepackage{pifont}
\usepackage{cite}
\usepackage{amsmath,amssymb,amsfonts}
\usepackage{algorithmic}
\usepackage{graphicx}
\usepackage{textcomp}
\usepackage{xcolor}
\def\BibTeX{{\rm B\kern-.05em{\sc i\kern-.025em b}\kern-.08em
    T\kern-.1667em\lower.7ex\hbox{E}\kern-.125emX}}
\begin{document}

\title{Reflection Generation for Composite Image Using Diffusion Model\\}

\author{
    Haonan Zhao, Qingyang Liu, Jiaxuan Chen, Li Niu$^*$\thanks{*Corresponding author.} \\
    Shanghai Jiao Tong University \\
    \{2zz-n-24, narumimaria, chenjiaxuan, ustcnewly\}@sjtu.edu.cn
}
\maketitle

\begin{abstract}
Image composition involves inserting a foreground object into the background while synthesizing environment-consistent effects such as shadows and reflections. Although shadow generation has been extensively studied, reflection generation remains largely underexplored. In this work, we focus on reflection generation. We inject the prior information of reflection placement and reflection appearance into foundation diffusion model. We also divide reflections into two types and adopt type-aware model design. To support training, we construct the first large-scale object reflection dataset DEROBA. Experiments demonstrate that our method generates reflections that are physically coherent and visually realistic, establishing a new benchmark for reflection generation. The Dataset and model are released at \href{https://github.com/bcmi/Object-Reflection-Generation-Dataset-DEROBA}{https://github.com/bcmi/Object-Reflection-Generation-Dataset-DEROBA}
\end{abstract}

\begin{IEEEkeywords}
image composition, reflection generation
\end{IEEEkeywords}

\section{Introduction}
\label{sec:intro}
\begin{figure}[t]
\centering
\includegraphics[width=0.75\linewidth]{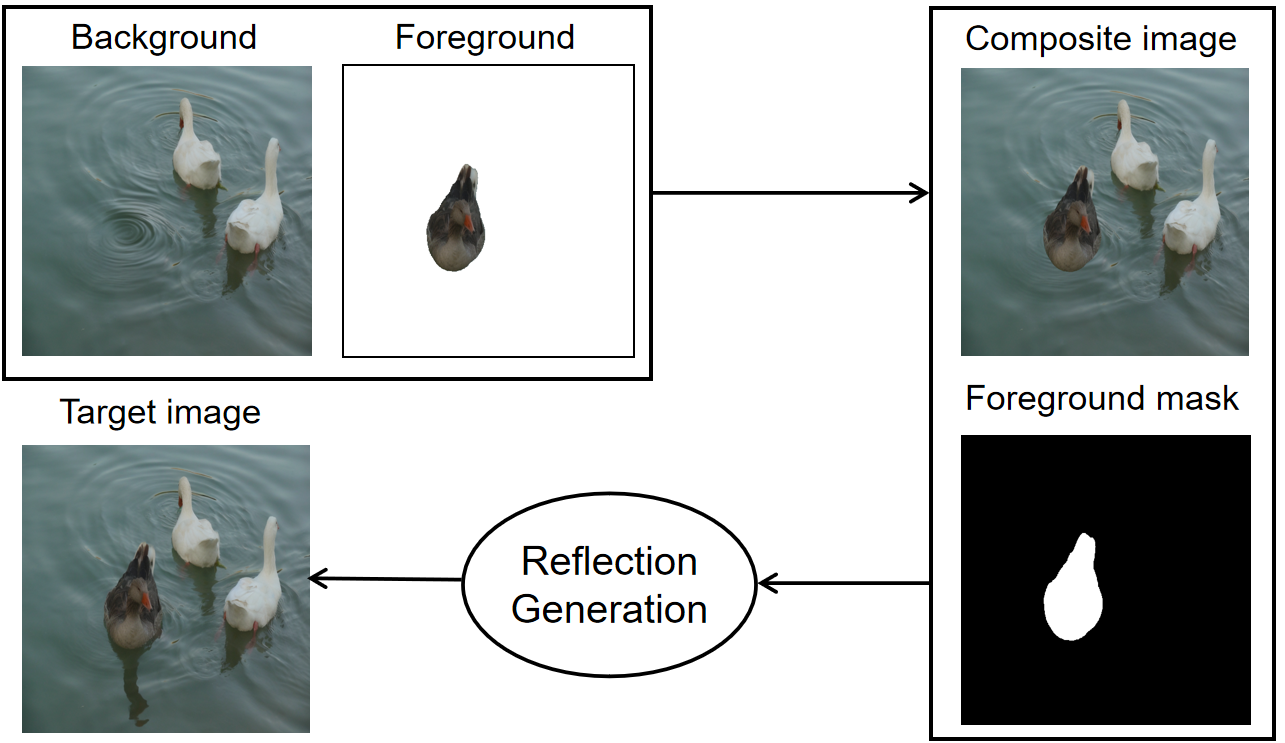}
   \caption{A composite image can be obtained by placing a foreground object onto the background. Reflection generation seeks to synthesize plausible reflection for the inserted object, thereby enhancing the overall realism and perceptual consistency of the composite image.}
\label{fig:fig_intro}
\end{figure}

The goal of image composition~\cite{niu2021making} is inserting a foreground object into the background. To make the composite image visually natural, it is essential to synthesize realistic shadows and reflections consistent with the surrounding environment. While shadow generation has been extensively explored~\cite{zhao2025shadow,liu2024shadow}, reflection generation is still largely underexplored. In this work, we focus on the task of reflection generation, which aims to generate realistic reflections for the specified foreground in a composite image as shown in the Fig.~\ref{fig:fig_intro}.

To leverage the strong generative prior of foundation diffusion models such as Stable Diffusion (SD)~\cite{rombach2022high}, we build our reflection generation model on top of SD-1.5. Following~\cite{zhang2023adding}, we add a ControlNet encoder with composite image and composite foreground mask being the input. 
To facilitate the reflection generation, we attempt to inject the placement information (location and scale) of foreground reflection into diffusion model. The location/scale of an object's reflection is related to the location/scale of the object itself in a certain pattern. For example, the reflection in the water is always below the object and the reflection scale is proportional to the object scale. Therefore, we employ an auxiliary encoder to predict the regression coefficients from foreground bounding box to reflection bounding box. The predicted regression coefficients lead to the predicted reflection bounding box, based on which we can obtain the reflection box mask and append it to the ControlNet encoder input. This mask provides the placement information of foreground reflection. 

Moreover, we observe that the reflection appearance is often the vertical flip of foreground object. We categorize reflections into two types: ``vertical flip" and ``others". 
The abovementioned auxiliary encoder jointly predicts the regression coefficients and reflection type.
For the type ``vertical flip", we crop the foreground object from composite image and use the vertically flipped foreground crop as reference image. The model could simply place the flipped foreground crop in the predicted reflection bounding box and adapt its style to reflection receiver (\emph{e.g.}, water).
For the type ``others", the task is much more complicated and we directly use the foreground crop as reference image, since the reflection appearance may be partially borrowed from the foreground appearance. We extract reference image features and inject them into diffusion model.
For two types of reflections, the model assumes different responsibilities. To distinguish such difference, we learn two reflection type embeddings which are injected into diffusion model along with reference features. 

\begin{figure*}[t]
\begin{center}
\includegraphics[width=0.65\linewidth]{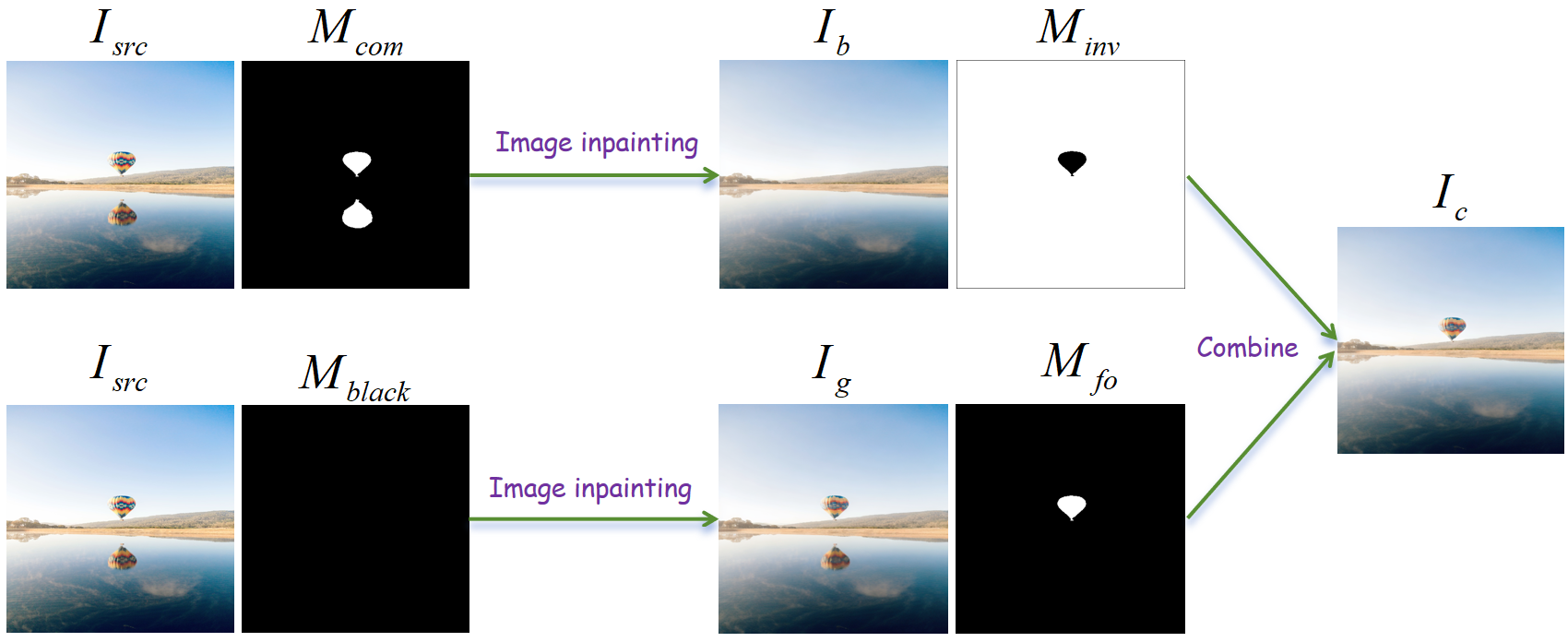}
\end{center}
   \caption{The pipeline of DEROBA dataset construction. Foreground objects are automatically detected and segmented, with their associated reflection masks manually annotated. Inpainting is performed on the source image $I_{src}$ within both foreground and reflection regions to generate a background image $I_{b}$. To correct color discrepancies, inpainting is again performed on $I_{src}$ with a black mask, leading to the ground-truth image $I_{g}$. Finally, $I_{b}$ and $I_{g}$ are combined to get the composite image $I_{c}$. The details can be found in Section~\ref{sec:dataset}.}
\label{fig:deroba_pipeline}
\end{figure*}

Training the reflection generation model requires pairs of reflection-free composite images and ground-truth reflection images. No existing dataset is tailored for this purpose, and the datasets \cite{li2023dreamedit,kim2025orida} for generic image composition lack explicit object–reflection associations. To address this gap, we construct the DE-reflected Reflection-Object Association (DEROBA) dataset, the first large-scale dataset specifically designed for reflection generation. We collect real-world object–reflection pairs from Pixabay, and automatically extract foreground masks using \cite{liu2024grounding,kirillov2023segment}. Corresponding reflection masks are then manually annotated, and reflection regions are inpainted to produce reflection-free composites paired with their ground-truth reflections.

We conduct experiments on our DEROBA to validate the effectiveness of our method. Both quantitative metrics and qualitative comparisons demonstrate that our model generates reflections that are realistic and well aligned with the surrounding scene. In summary, our contributions are: 1) We construct DEROBA, the first large-scale dataset specifically designed for learning object–reflection associations; 2) We propose to use the prior information of reflection placement and reflection appearance, as well as reflection type-aware model design;  3) We validate our framework on  DEROBA dataset, demonstrating its clear advantages in producing physically consistent and visually compelling reflections.
\begin{figure*}[t]
\begin{center}
\includegraphics[width=0.82\linewidth]{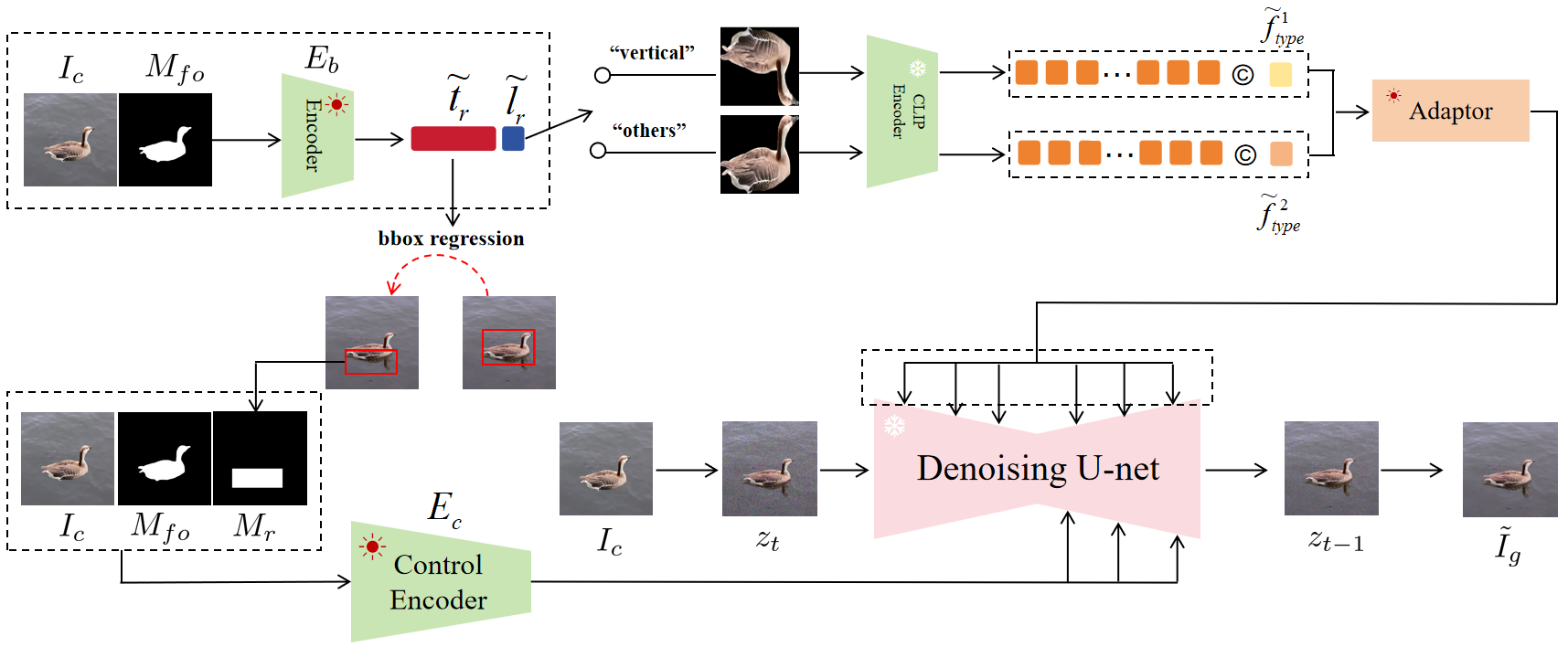}
\end{center}
   \caption{The illustration of our method. Besides denoising U-Net and ControlNet encoder $E_{c}$,  we introduce an auxiliary encoder $E_{b}$ to predict the bounding box regression coefficients $\tilde{l}_r$ and reflection type $\tilde{t}_r$. According to the predicted reflection type, we extract the corresponding reference features and use the corresponding reflection type embedding.}
\label{fig:model}
\end{figure*}
\section{Related Work}
\label{sec:formatting}

\subsection{Image Composition}
Early works on image composition are divided into different subfields like image harmonization~\cite{cong2020dovenet,cong2022high,tao2024diverse}, shadow generation~\cite{zhao2025shadow,liu2024shadow}, and so on. 
In recent years, abundant generative image composition  works~\cite{winter2024objectmate,song2024imprint,wang2024primecomposer,yuan2023customnet,yu2025omnipaint,wang2025unicombine} have emerged to seamlessly integrate a given foreground object into a target image. 
The early diffusion-based works~\cite{yang2023paint,Song_2023_CVPR} leverage conditional diffusion model and propose self-supervised framework.
ControlCom~\cite{zhang2023controlcom} introduces a controllable image composition method that unifies four tasks: image blending, image harmonization, view synthesis, and generative composition.  
CareCom~\cite{chen2026carecom} proposes to hallucinate reference features matching the background. 
More recently, Diffusion Transformers (DiT) have also been applied to image composition tasks~\cite{wang2025unicombine,zhang2025easycontrol,song2025insert}. For example, EasyControl~\cite{zhang2025easycontrol} introduces a unified and highly efficient conditional DiT framework featuring plug-and-play condition injection, position-aware training, and causal attention with KV caching. Insert Anything~\cite{song2025insert} further proposes a DiT-based reference insertion framework that leverages multimodal attention and in-context editing. However, these methods primarily focus on general composition rather than explicit reflection generation, leaving the challenge of high-quality reflection synthesis underexplored.

\subsection{Reflection Generation}
Recently, some works focus on generating highly realistic and plausible mirror reflections. For example, MirrorFusion~\cite{dhiman2025reflecting} introduces SynMirror, a large-scale synthetic dataset for generating mirror reflections using depth-conditioned inpainting. Similarly, MirrorVerse~\cite{dhiman2025mirrorverse} enhances mirror reflection generation by employing augmentations such as random object positioning, rotations, and grounding. These methods are highly effective in creating  geometrically accurate mirror reflections.
In contrast, our work focuses on a wide range of reflections on diverse surfaces.  

Similar to our work,  \cite{canet2025thinking} introduces unconstrained image composition, which can generate reflection outside the foreground bounding  box. However, the oversimplified approach to obtain reflection masks during dataset construction limits the data quality and may mislead the model learning. Moreover, its dataset and model are not publicly available. 
\section{Reflection Dataset Construction}
\label{sec:dataset}
The pipeline of constructing our DEROBA dataset is illustrated in Fig.~\ref{fig:deroba_pipeline}. First, we collect a large number of high-resolution images from various indoor and urban scenes, and use off-the-shelf tools (BLIP2~\cite{li2023blip}, Grounding DINO~\cite{liu2024grounding} and SAM~\cite{kirillov2023segment}) combined with manual annotation to get the foreground mask $M_{fo}$ and reflection mask $M_{r}$.

Next, we combine $M_{fo}$ and $M_{r}$ to get $M_{com}$. We then use the image inpainting model \cite{zhuang2024task} to inpaint the original real image $I_{src}$ within the region $M_{com}$, generating the background image $I_{b}$. However, the inpainted background image always exhibits certain color discrepancy compared with $I_{src}$. Directly using $I_{src}$ as the ground-truth would lead the model to learn incorrect color discrepancy. Therefore, we  additionally inpaint $I_{src}$ using completely black mask $M_{black}$, aiming to produce the ground-truth image $I_{g}$ with no color discrepancy from $I_{b}$.

Finally, we combine $I_{b}$ and $I_{g}$ using the foreground mask $M_{fo}$ and its inverse mask $M_{inv}$ to generate the composite image $I_{c}$. $I_{c}$, $M_{fo}$, and $I_{g}$, form a training tuple that can be used for model training.

Note that after inpainting, we manually filter out the tuples with low-quality reflection or object masks, as well as the tuples with noticeable artifacts or unnatural inpainted areas, to ensure the high quality of the dataset. Example images of the dataset and more detailed analysis of the construction process are presented in the supplementary material.
\begin{figure*}[t]
\centering
\includegraphics[width=0.9\linewidth]{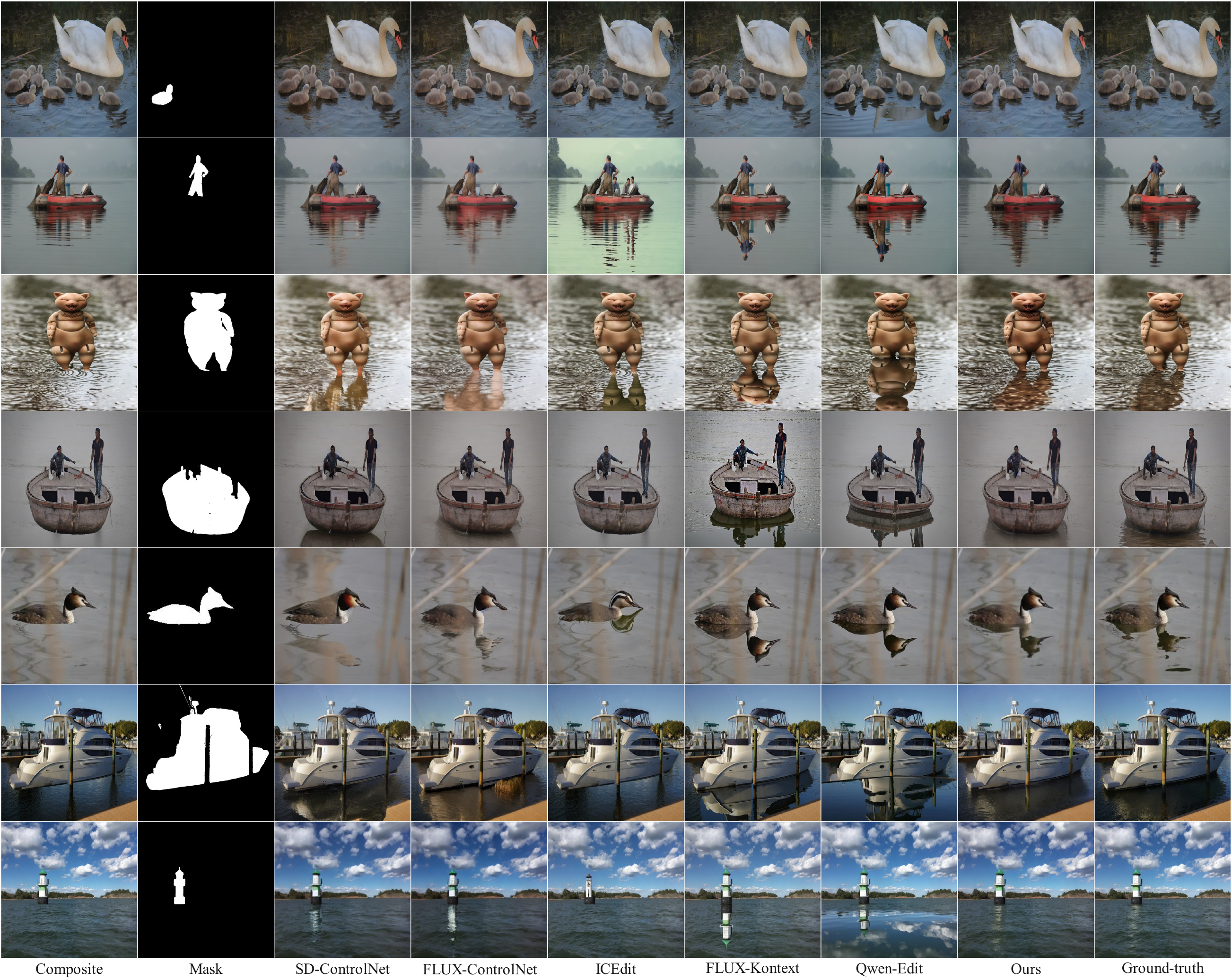}
\caption{Visual comparison of different methods on DEROBA dataset. From left to right are input composite image, foreground mask, results of SD-ControlNet, FLUX-ControlNet, ICEdit, FLUX-Kontext, Qwen-Edit, our method, and ground-truth.}
\label{fig:vis_deroba_main}
\end{figure*}
\section{Method}
\label{sec:method}

The overall pipeline of our method is shown in Fig.~\r ef{fig:model}. Given the composite image $I_{c}$ and foreground mask $M_{fo}$, our goal is to create realistic reflection for the foreground object. The model is built upon Stable Diffusion~\cite{rombach2022high} and ControlNet~\cite{zhang2023adding}.
Besides, we introduce an auxiliary encoder $E_{b}$. On one hand, it predicts the reflection bounding box, based on which we can obtain the reflection box mask and append it to ControlNet input.  This will be introduced in Section~\ref{sec:box}. 
On the other hand, it predicts the reflection type (``vertical" or ``others"),
Based on the predicted type, we concatenate the corresponding reflection type embeddings and reference image features, which are injected into diffusion model via cross-attention. This will be introduced in Section~\ref{sec:clip}.

\subsection{Reflection Type and Bounding Box Prediction}
\label{sec:box}

The input of auxiliary encoder $E_{b}$ includes the composite image $I_{c}$ and foreground mask $M_{fo}$. The output of $E_{b}$ is a 6-dimensional vector, where the first dimension $\tilde{l}_{r}$ is the predicted reflection type (``vertical" or ``other"), and the remaining five dimensions are the regression coefficients $\tilde{t}_{r}$ for the rotated reflection bounding box. 

For the first dimension, with $l_{r}$ denoting the ground-truth label of reflection type, we supervise the predicted reflection type $\tilde{l}_{r}$ using cross-entropy loss: $\mathcal{L}_{cls}=\mathcal{L}_{ce}(\tilde{l}_{r}, l_{r})$.

For the remaining five dimensions, following the convention in the field of object detection, we use $B_{o}=(x_{o}, y_{o}, w_{o}, h_{o}, \theta_{o})$ to represent the rotated bounding box coordinates of the foreground object, and $B_{r}=(x_{r}, y_{r}, w_{r}, h_{r}, \theta_{r})$ to represent the box coordinates of object reflection. The relationship between these two can be expressed using the regression coefficient $t_{r}=(t_{x}^{r}, t_{y}^{r}, t_{w}^{r}, t_{h}^{r}, t_{\theta}^{r})$ by
\begin{equation}
\begin{cases}
  t_x^{r} = (x_{r} - x_{o}) / w_{o}, \\
  t_y^{r} = (y_{r} - y_{o}) / h_{o}, \\
  t_{w}^{r} = \ln (w_{r} / w_{o}), \\
  t_h^{r} = \ln (h_{r} / h_{o}), \\
  t_{\theta}^{r} = (\theta_{r} - \theta_{o}) \cdot \pi / 180.
\end{cases}
\label{eq:box_r}
\end{equation}

We aim to predict the regression coefficient $\tilde{t}_{r}$. Then, the reflection bounding box $\tilde{B}_{r}$ can be derived from $B_{o}$ and $\tilde{t}_{r}$. We supervise the predicted reflection bounding box by
\begin{equation}
\begin{aligned}
\mathcal{L}_{rbbox} & = \mathcal{L}_{kfiou}(\tilde{B}_{r}, B_{r}) = e^{1-KFIoU(\tilde{B}_{r}, B_{r})}.
\end{aligned}
\end{equation}

Given the predicted reflection bounding box $\tilde{B}_{r}$, we transform it to binary mask $M_{r}$, and concatenate it with $\{I_{c}, M_{fo}\}$. The concatenation is fed into the ControlNet encoder $E_{c}$.

We train the auxiliary encoder $E_{b}$ with the loss $\mathcal{L}_{en}  = \mathcal{L}_{cls} + \mathcal{L}_{rbbox}$.

\subsection{Reflection Type-Aware Diffusion Model}
\label{sec:clip}

\subsubsection{Reflection type-aware reference images}
We first crop the foreground object $F_{g}$ from the composite image $I_{c}$. Then, based on the predicted reflection type, we decide how to process the foreground crop $F_{g}$. For the type ``vertical", $F_{g}$ is flipped vertically, since the reflection appearance resembles the vertically flipped foreground. For the type ``others", we leave $F_{g}$ as it is, as the reflection appearance may
be partially borrowed from the foreground appearance. The processed $F_{g}$ is used as reference image and passed through the CLIP encoder to extract reference features $f_{ref}$. 

\subsubsection{Reflection type embeddings}
 
For two types of reflections, the model assumes different responsibilities. For the type ``vertical", the model could simply put the flipped foreground
crop in the reflection bounding box and submerge it into the reflection receiver. For the type ``others", the task for the model could be much more challenging. 
To distinguish such difference, we define two learnable reflection type embeddings $\{f_{type}^{k}|_{k=1}^2\}$, which are jointly optimized with the model. 

The reference features $f_{ref}$ and reflection type embeddings $\{f_{type}^{k}|_{k=1}^2\}$ are fed into an adapter to yield $\tilde{f}_{ref}$ and $\{\tilde{f}_{type}^{k}|_{k=1}^2\}$ respectively. They are subsequently injected into the denoising U-Net via cross-attention. 

Specifically, we employed the decoupled cross-attention structure for  $\tilde{f}_{ref}$ and $\{\tilde{f}_{type}^{k}|_{k=1}^2\}$ following \cite{ye2023ip}. Given image query features $f_{q}$ in the denoising U-Net, the output of decoupled cross-attention can be represented as
\begin{equation} 
\begin{aligned}
    f_{out} &= \text{Attention}(Q, K, V) + \text{Attention}(Q, K', V'),
\end{aligned}
\label{eq:att_r}
\end{equation}
where $Q=f_{q} W_{q}$, $K=\tilde{f}_{type} W_{k}$, $V=\tilde{f}_{type} W_{v}$, $K'=\tilde{f}_{ref} W'_{k}$, $V'=\tilde{f}_{ref} W'_{v}$. Here, $W_{q}$, $W_{k}$, $W_{v}$, $W'_{k}$, and  $W'_{v}$ are the weight matrices of the linear projection layers. $Q$ is the derived from $f_{q}$. $K$ (\emph{resp.}, $K'$) , $V$ (\emph{resp.}, $V'$) are derived from $\tilde{f}_{type}$ (\emph{resp.}, $\tilde{f}_{ref}$). 

During training, we add noise to the ground-truth $I_{g}$ and feed noisy latent $z_{t}$ into diffusion model to obtain the predicted image $\tilde{I}_{g}$. Given the predicted reflection box mask $M_r$, the reference features $f_{ref}$, and reflection type embeddings $f_{type}^{k}$, we supervise the training process by
\begin{equation} 
	\mathcal{L}_{dm} = \mathbb{E}_{t,  \epsilon \sim \mathcal{N}(0, 1) }\Big[ \Vert \epsilon - \epsilon_{\phi}(z_{t}, t, M_r, f_{ref}, f_{type})) \Vert_{2}^{2}\Big],
\end{equation}
where $t$ is current timestep, $\epsilon$ is the added noise and $\epsilon_{\phi}$ parameterized by $\phi$ predicts the added noise. 

During inference, the noisy latent input is generated by adding noise to the composite image $I_{c}$ to enhance the retention of original details.

\begin{table}[t]
\caption{The results of different methods on DEROBA test set. The best results are highlighted in boldface.}
\centering
\vspace{0pt}
\centering
\resizebox{0.8\linewidth}{!}{
\begin{tabular}{c|cccc}
    \toprule
    Method
    &GR $\downarrow$ &LR $\downarrow$ &GS $\uparrow$  &LS $\uparrow$\cr
    \cmidrule(r){1-1} 
    \cmidrule(r){2-5}  
    SD-ControlNet~\cite{rombach2022high} & 15.726 & 57.110 & 0.903 & 0.103 \cr
    FLUX-ControlNet~\cite{zhang2023adding} &15.451 & 57.882 & 0.898 & 0.115 \cr
    ICEdit~\cite{zhang2025context} & 19.464 &59.643 & 0.862& 0.095\cr
    FLUX-Kontext~\cite{batifol2025flux} &13.839 &55.231 &0.913 & 0.134\cr
    Qwen-Edit~\cite{wu2025qwen} &12.390 &54.797 &0.916 & 0.129\cr
    Ours&\textbf{11.522} &\textbf{53.430} & \textbf{0.923}& \textbf{0.144} \cr
    \bottomrule
\end{tabular}
}
\vfill
\vspace{3pt}
\label{tab:DESOBA_DEROBA}
\end{table}

\begin{figure*}[ht]
\centering
\includegraphics[width=0.85\linewidth]{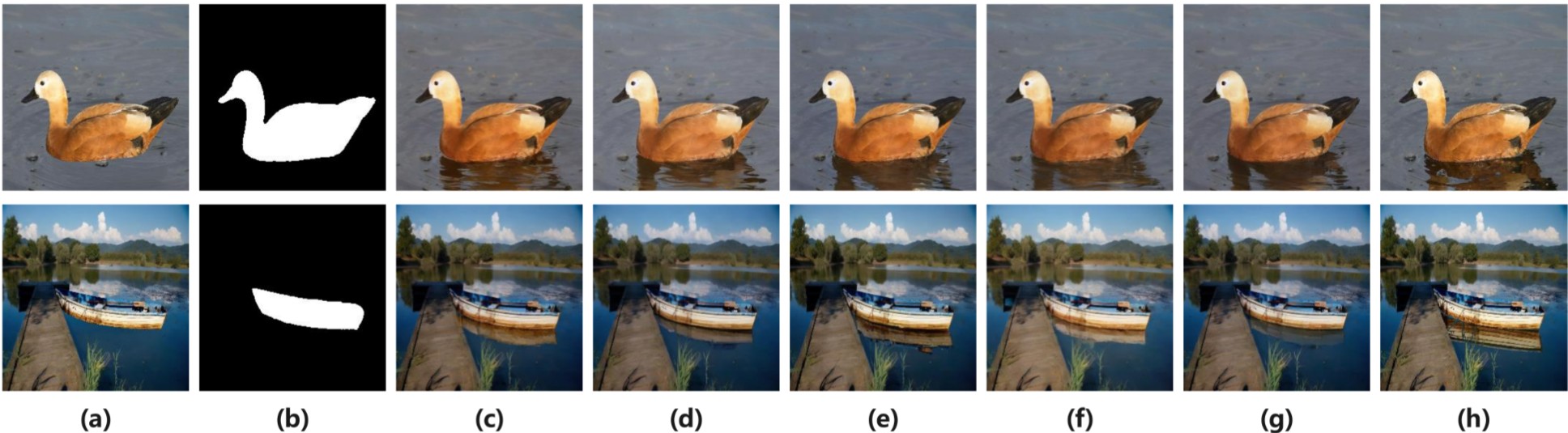}
\caption{Multiple results for one test image on DEROBA test set. From left to right are input composite image (a), foreground object mask
(b), results of our method using different random seeds (c)-(g) and ground-truth (h).}
\label{fig:seed}
\end{figure*}

\section{Experiments}
\label{sec:experiments}
\subsection{Datasets and Evaluation Metrics}

We split our DEROBA dataset into 20387 training tuples and 629 test tuples. For performance evaluation on the test set, we calculate the difference between the generated image and ground-truth image. Specifically, local RMSE (LR) and Local SSIM (LS) focus on the reflection region, while Global RMSE (GR) and Global SSIM (GS) apply to the entire image. 

\subsection{Implementation Details}
\label{sec:implementation}
We develop our method with PyTorch 2.4.0. We employ ResNet18 \cite{he2016deep} as the encoder $E_b$. We obtain image-specific reflection features using CLIP-ViT-B-32~\cite{radford2021learning}. The adaptor structure is a simple linear mapping layer followed by a LayerNorm. We train ControlNet and adaptor using the Adam optimizer with a constant learning rate of $1e^{-5}$ over 10 epochs on four NVIDIA RTX A6000 GPUs. 

\subsection{Baselines}
Since there is currently no open-source model directly used for reflection generation, we train ControlNet on DEROBA by using the composite image and foreground mask as input. We train two versions of ControlNet: SD-ControlNet and FLUX-ControlNet, by using SD-1.5 and FLUX.1-dev as base model. 

We also compare with two SOTA open-source image editing models that can generate reflection, ICEdit~\cite{zhang2025context}, FLUX-Kontext~\cite{batifol2025flux}, and Qwen-Edit~\cite{wu2025qwen} using ``add reflection to [foreground category]" as instruction prompt to generate foreground reflection. We finetune these models with LoRA on our DEROBA dataset. 

\subsection{Evaluation on DEROBA Dataset}
The quantitative results  on DEROBA are summarized in Table \ref{tab:DESOBA_DEROBA}. Our method achieves the lowest GRMSE, LRMSE and highest GSSIM, LSSIM, which demonstrates that our method could generate reflections closest to the ground-truth.

The visual comparison of reflection generation are shown in Fig.~\ref{fig:vis_deroba_main}. Our model produces reflections with more realistic shapes and finer details compared to other baselines, while also maximizing the background preservation. Even in complex scenes with multiple objects that occlude each other (\emph{e.g.}, the third row), our method can still generate realistic reflections. 
ControlNet and ICEdit often fail to generate complete reflections. 
The reflections generated by FLUX-Kontext and Qwen-Edit are prone to be too clear, ignoring the ripples on the water surface, and thus looking very unrealistic.
More visual reflections results can be found in supplementary.

\begin{table}[t]
  \caption{Ablation studies of our method on DEROBA test set. }
  \begin{center}
  \resizebox{0.9\linewidth}{!}{
  \begin{tabular}{c|cccc|cccccc}
      \toprule[0.8pt]
      Row & Base & Box & Features & Type &GR $\downarrow$ &LR $\downarrow$  &GS$\uparrow$&LS$\uparrow$ \cr
      \hline
    1 & + & - & - & - & 15.726 & 57.110 & 0.903 & 0.103 \cr
    2 & + & + & - & - &  13.917 & 54.873 & 0.904 & 0.131 \cr
    3 & + & - & + & - & 12.275 & 55.354 & 0.915 & 0.127 \cr
    4 & + & + & + & - & 11.739 & 53.922 & 0.918 & 0.139 \cr
    5 & + & + & + & + & 11.522 & 53.430 & 0.923 & 0.144 \cr
    \bottomrule
    \end{tabular}}
    \end{center}
    \label{tab:ablation_deroba}
\end{table}

\subsection{Multiple Results for One Test Image}
\label{sec:seed}

Since diffusion model has stochastic property, we show five results of our method for each test image by using different random seeds in  Fig.~\ref{fig:seed}. It can be seen that our method performs stably and generates overall satisfactory results. 

\subsection{Ablation Studies}
\label{sec:vis_ablated_methods}
We conduct ablation studies on our method to justify the necessity of each component in  Table~\ref{tab:ablation_deroba}. 
In row 1, we report the results of basic ControlNet model without reflection bounding box prediction or reference features injection. Then, we report the results with reflection bounding box prediction in row 2, with reference features injection in row 3, with both in row 4. It can be seen that both components contribute to the performance and jointly using them can further improve the results. Finally, we introduce reflection type embedding and concatenate it with reference features. As reported in row 5, distinguishing two types of tasks leads to comprehensive improvement over all metrics.

\subsection{Limitations Discussion}
\label{sec:limitation}
\begin{figure}[t]
\centering
\includegraphics[width=0.85\linewidth]{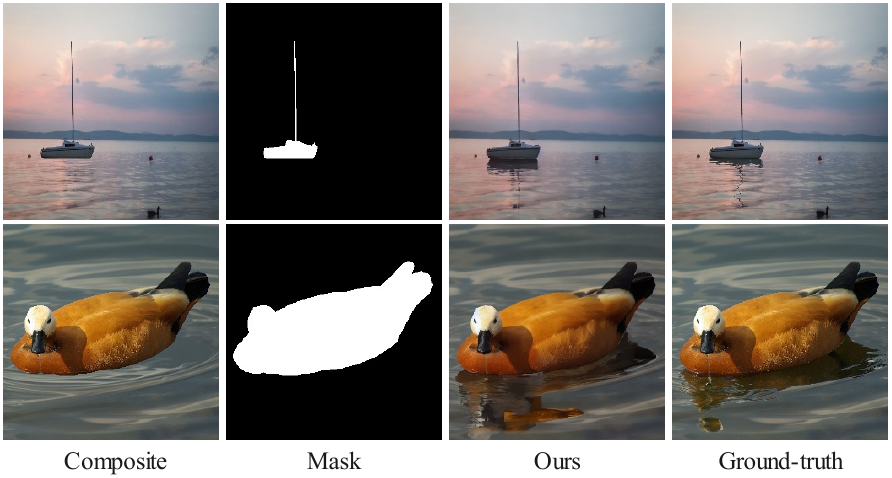}
\caption{Visual results of failure cases. From left to right, we show composite image, foreground object mask, our result, and ground-truth.}
\label{fig:failure_case}
\end{figure}

Our method generally produces reasonable reflections. However, the results may not be satisfactory in certain complex scenes. Fig.~\ref{fig:failure_case} shows some failure cases. For instance, our model struggles in the complex fluid scenes, as well as the scenes where the object is intact but the reflection has gaps.

\section{Conclusion}
In this work, we have constructed the first large-scale dataset DEROBA specifically designed for reflection generation task. 
We have also proposed a novel reflection generation method with the prior of reflection placement and appearance. The experimental results on DEROBA demonstrate the effectiveness of our proposed method.

\section*{Acknowledgment}

The work was supported by the National Natural Science Foundation of China (Grant No. 62471287).

\bibliographystyle{IEEEbib}
\bibliography{icme2026references}

\end{document}


\title{Supplementary for Reflection Generation for Composite Image Using Diffusion Model}

\author{
    Haonan Zhao, Qingyang Liu, Jiaxuan Chen, Li Niu$^*$\thanks{*Corresponding author.} \\
    Shanghai Jiao Tong University \\
    \{2zz-n-24, narumimaria, chenjiaxuan, ustcnewly\}@sjtu.edu.cn
}

\maketitle

In this supplementary, we provide additional materials to support our main submission. In Section~\ref{sec:deroba}, we introduce more details about the construction process of DEROBA dataset. In Section~\ref{sec:statistics}, we provide example images and more statistics of our DEROBA dataset. In Section~\ref{sec:ablation}, we present the visual results of ablation studies, demonstrating the contribution of each component in our model. In Section~\ref{sec:encoder}, we further discuss reflection type classification and bounding box prediction, including prediction accuracy and visualization results. In Section~\ref{sec:more_result}, we present more visual comparison results with baselines. In Section~\ref{sec:time}, we report the comparison of time complexity.

\section{Detailed Reflection Dataset Construction}
\label{sec:deroba}
The pipeline of DEROBA construction is shown in Fig. 2 in the main paper. We collect a large number of high-resolution and license-free images from Pixabay, retaining only those that contain at least one object-reflection pair. Initially, we use BLIP2~\cite{li2023blip} to generate descriptive captions for each image (\emph{e.g.}, ``Photo of a goose swimming in a pond”). The caption is then input into Grounding DINO~\cite{liu2024grounding} to detect potential foreground objects and obtain their bounding boxes.

We then use the predicted foreground object bounding boxes (\emph{e.g.}, ``goose” and ``pond”) as input for SAM~\cite{kirillov2023segment}, which generates precise segmentation masks $\{M_{o,n}|_{n=1}^N\}$ for the detected objects. To ensure high data quality, professional human annotators manually filter out object masks that do not contain reflections  (\emph{e.g.}, ``pond”) while retaining those that contain reflections $\{M_{o,n}|_{n=1}^K\}$ (\emph{e.g.}, ``goose”). Human annotators also carefully annotate the corresponding reflection masks $\{M_{r,n}|_{n=1}^K\}$ based on the real image $I_{r}$ and the object masks$\{M_{o,n}|_{n=1}^K\}$. Finally, we obtained 21,016 object-reflection pairs $(M_{o,k}, M_{r,k})$, in which $M_{o,k}$ (\emph{resp.}, $M_{r,k}$) is the mask of the $k$-th object (\emph{resp.}, reflection). 

Then, for each object-reflection pair, we use the inpainting pipeline (see Fig. 2 in the main paper) to construct composite image $I_c$ and ground-truth image $I_g$. We filter out those inpainting failure cases and obtain 21,016 data tuples $\{I_{c}, M_{fo}, M_{r}, I_{g}\}$, in which 
$M_{fo}$ is foreground object mask and $M_{r}$ is foreground reflection mask. We split all the tuples into 20,387 tuples for training and 629 tuples for testing.

\begin{figure}[t]
\centering
\begin{minipage}{0.47\linewidth}
    \centering
    \includegraphics[width=\linewidth]{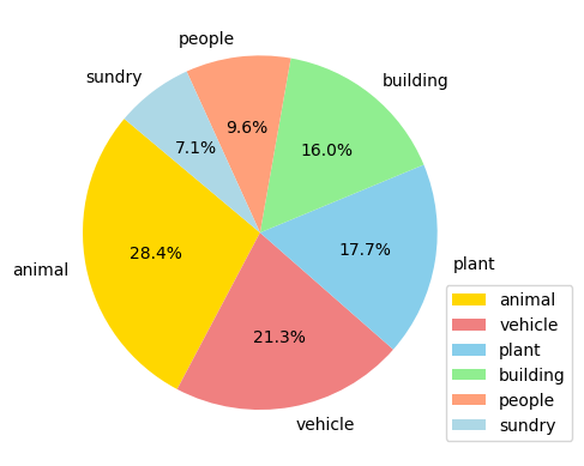}
    \caption{The super-category distribution of foreground object in DEROBA dataset.}
    \label{fig:category}
\end{minipage}\hfill
\begin{minipage}{0.47\linewidth}
    \centering
    \includegraphics[width=\linewidth]{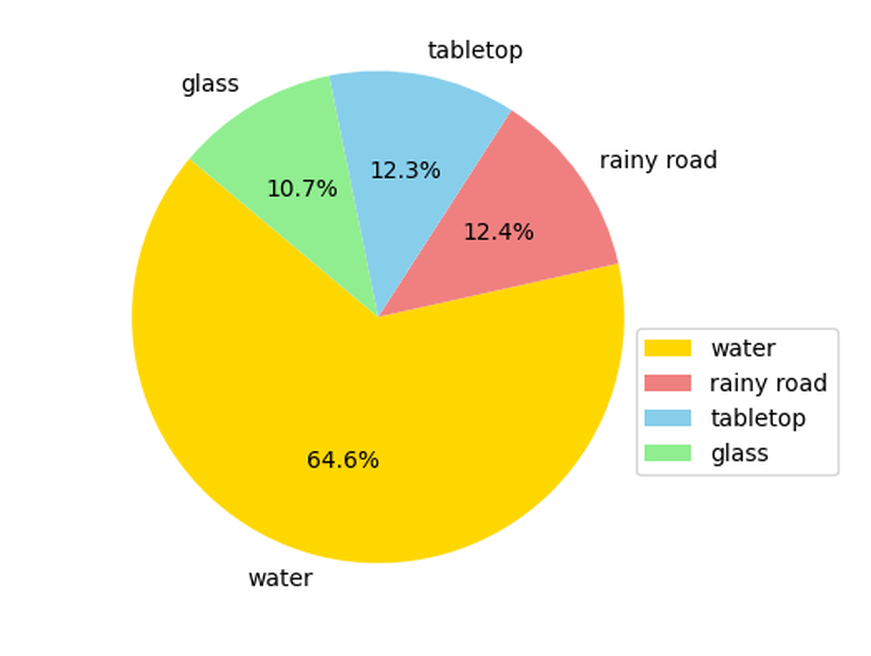}
    \caption{The super-category distribution of reflective surfaces in DEROBA dataset.}
    \label{fig:surfaces}
\end{minipage}
\end{figure}

\begin{figure}[t]
\centering
\includegraphics[width=0.85\linewidth]{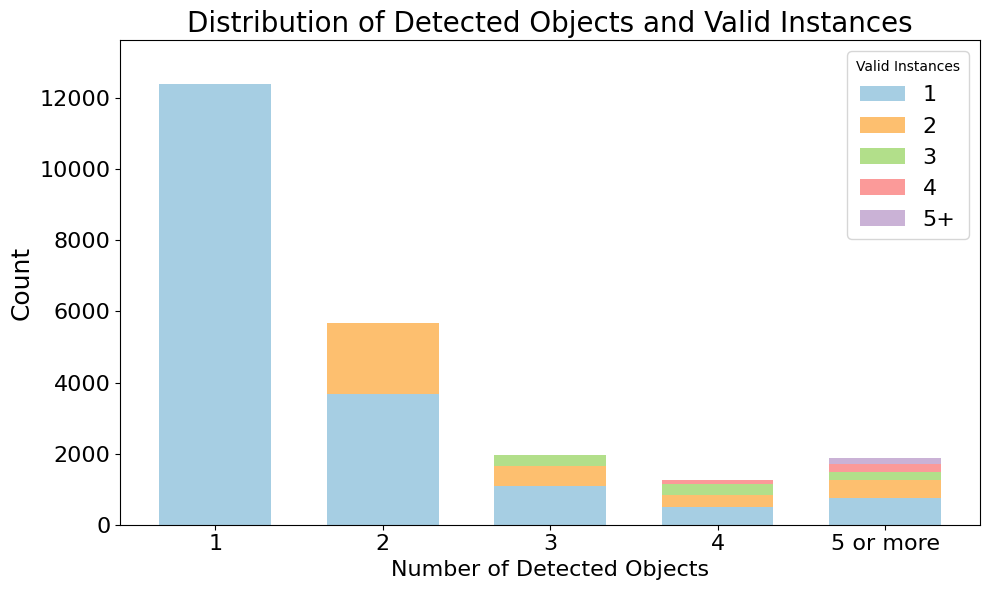}
\caption{The distribution of images with different numbers of detected foreground object in DEROBA dataset. Each bar further contains the distribution of images with different numbers of valid instances.}
\label{fig:histograms}
\end{figure}

\begin{figure*}[t]
\centering
\includegraphics[width=0.99\linewidth]{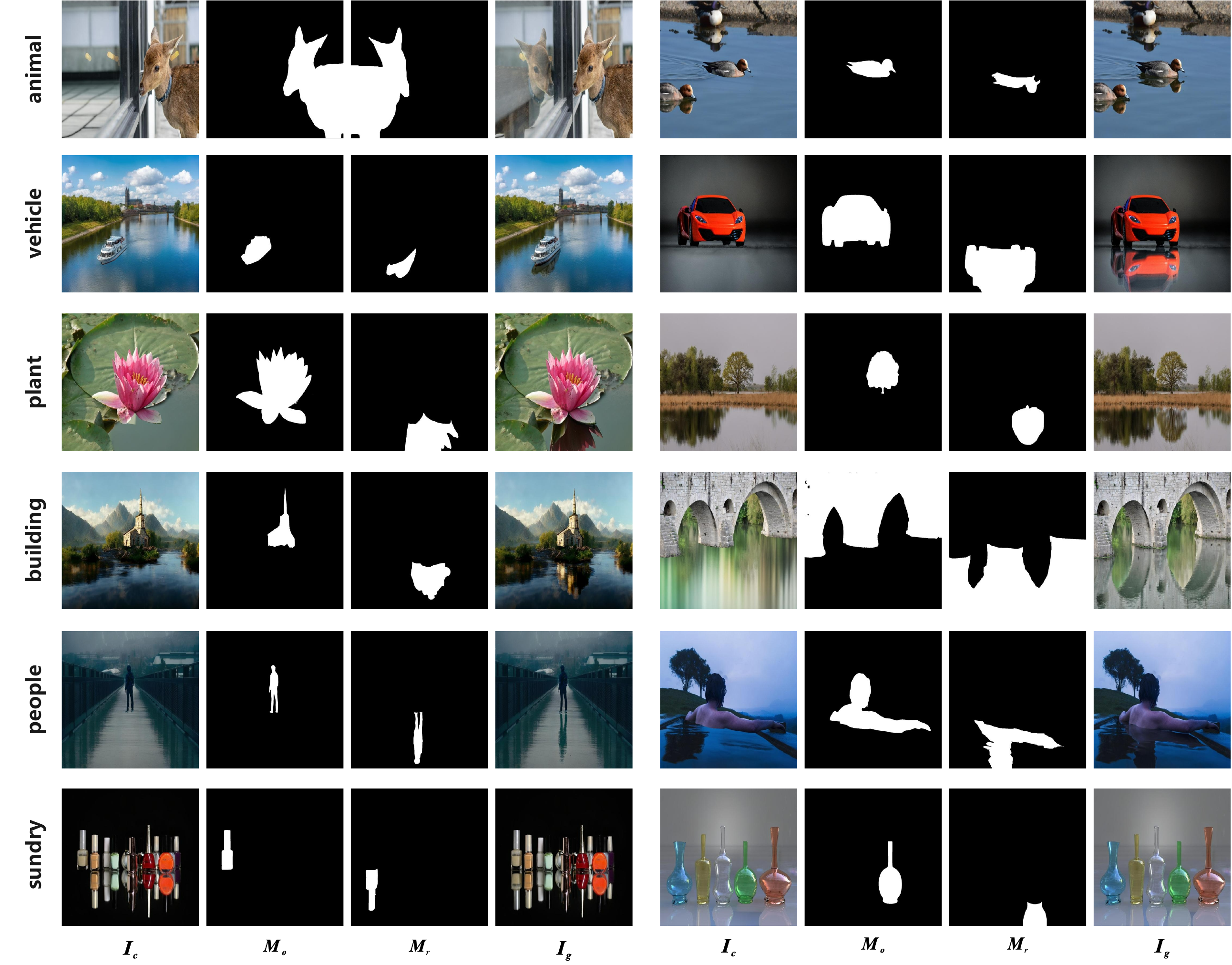}
\caption{Examples of DEROBA dataset in six super-categories (``animal", ``vehicle",  ``plant", ``building", ``people", ``sundry"). For each super-category, we present two tuples of composite image $I_{c}$,  foreground object mask $M_{fo}$, foreground reflection mask $M_{r}$, and ground-truth target image $I_{g}$.}
\label{fig:datasetexample}
\end{figure*}

\section{More Statistics of Our DEROBA Dataset} \label{sec:statistics}

We divide the objects in DEROBA dataset into  six super-categories: animal, vehicle, plant, building, people, sundry. 
We provide some examples of six super-categories from our DEROBA dataset in Fig.~\ref{fig:datasetexample}. For each super-category, we show two tuples in the form of $\{I_{c}, M_{fo}, M_{r}, I_{g}\}$, in which $I_{c}$ is composite image, $M_{fo}$ is foreground object mask, $M_{r}$ is foreground reflection mask, and $I_{g}$ is ground-truth target image. 

Moreover, we plot the super-category distribution of foreground objects in our DEROBA dataset in Fig.~\ref{fig:category}. It can be seen that our DEROBA dataset covers a diversity of categories, in which ``animal" is the dominant super-category. In addition, we also calculate the proportion of different reflective surfaces in our dataset. As shown in Fig.~\ref{fig:surfaces}, the reflective media in our dataset include water, rainy road, tabletop, and glass, covering most reflective scenes in the real world.

Recall that we use Grounding DINO~\cite{liu2024grounding} to detect potential foreground objects, which are referred to as detected instances. After manually filtering the objects without reflections and inpainting failure cases, we refer to the remaining instances as valid instances. Next, we summarize the statistics of detected instances and valid instances in our DEROBA dataset. Our DEROBA dataset has in total $16,791$ images with $21,016$ tuples. In Fig.~\ref{fig:histograms}, we first plot the distribution of images with different numbers of detected instances, based on which most images have fewer than $5$ detected instances. Among the images with specific number of detected instances, we further plot the distribution of images with different numbers of valid instances. Note that all images in our dataset have at least one valid instance, so $12,388$ images with one detected instance all have one valid instance. The images with more than one detected instance have different numbers of valid instances.

\section{Visualization of Ablation Studies}
\label{sec:ablation}
\begin{figure*}[t]
\centering
\includegraphics[width=0.9\linewidth]{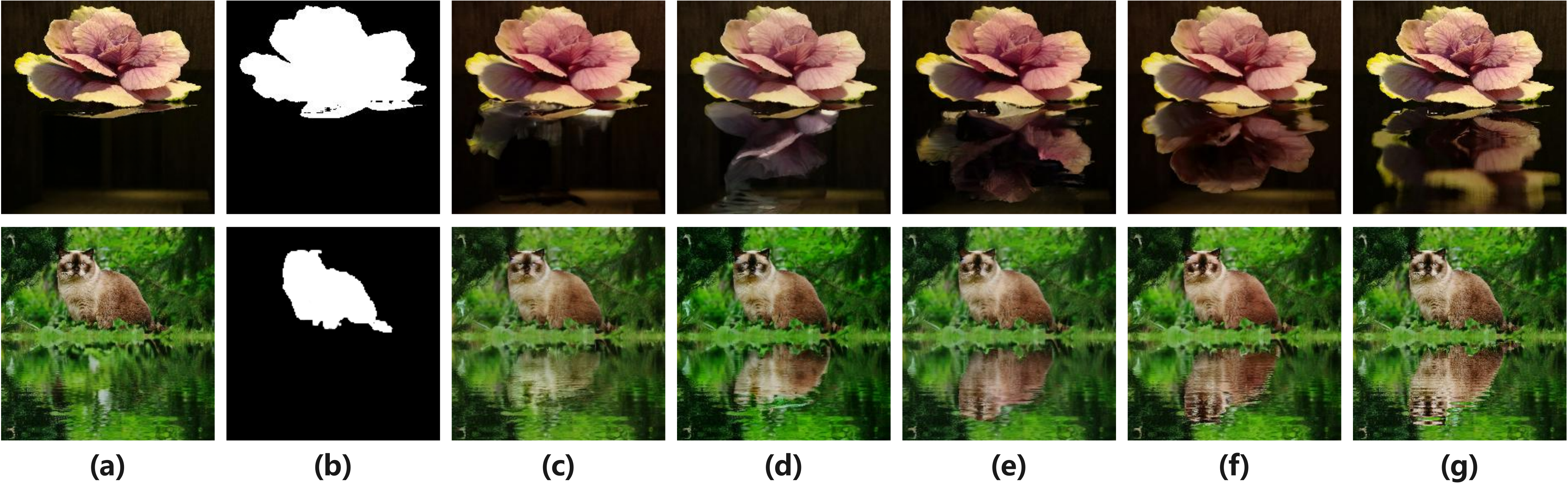}
\caption{The visual results of our ablated versions. From left to right are input composite image (a), foreground mask (b), basic ControlNet (c), only using reference features (d), only using reflection bounding box (e), full version (f), and ground-truth (g).}
\label{fig:ab_mw}
\end{figure*}

in Fig.~\ref{fig:ab_mw}, we provide the visual results of ablation studies (see Table 2 in the main paper), to further validate the effectiveness of each module. 
We show the results of basic ControlNet (row 1 in Table 2), only using reference features (row 3 in Table 2), only using reflection bounding box (row 2 in Table 2), and full version (row 5 in Table 2). 
It can be seen that basic ControlNet (c) achieves poor results. 
Without using reflection bounding box (d), the placement of generated reflection is not very accurate. Without using reference features (e), the generated reflection lacks some appearance details. Our full version (f) can generate reflections with precise placement and rich details, which are closer to the ground-truth (g).

\begin{table}[t]
  \begin{center}
  \caption{The classification accuracy of predicted reflection types.}
  \begin{tabular}{c|cc}
      \toprule[0.8pt]
      Metric & Training set & Test set\cr
      \midrule
      Acc $\uparrow$ & 0.962 & 0.945 \\
    \bottomrule
    \end{tabular}
    \end{center}
    \label{tab:cls}
\end{table}

\begin{table}[t]
  \begin{center}
  \caption{Quantitative results of the rotated bounding box regression. ``Upper" denotes the upper bound of KFIoU.}
  \begin{tabular}{c|ccc}
      \toprule[0.8pt]
      Metric & Training set & Test set & Upper \cr
      \midrule
    KFIoU $\uparrow$ & 0.263 & 0.246 & 0.333 \cr
    \bottomrule
    \end{tabular}
    \end{center}
    \label{tab:bbx}
\end{table}

\begin{figure}[t]
\centering
\includegraphics[width=0.99\linewidth]{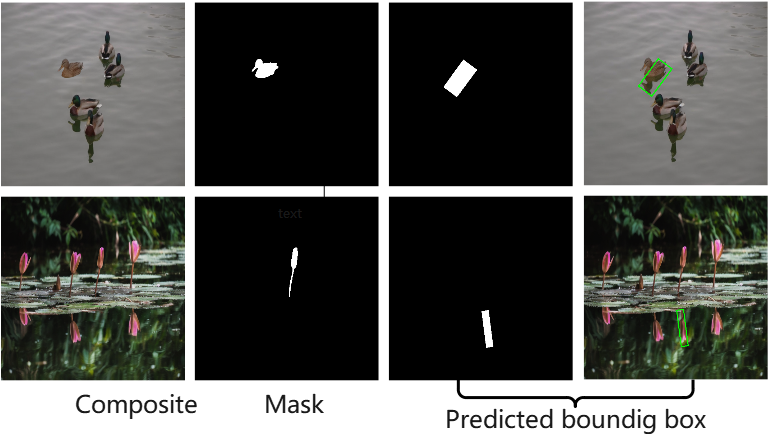}
\caption{Visualization results of rotated bounding box regression. Mask means foreground mask. In the third column, we present the predicted bounding box, while in the fourth column, we overlay it on the ground-truth image.}
\label{fig:box_supp}
\end{figure} 

\begin{figure*}[t]
\centering
\includegraphics[width=0.99\linewidth]{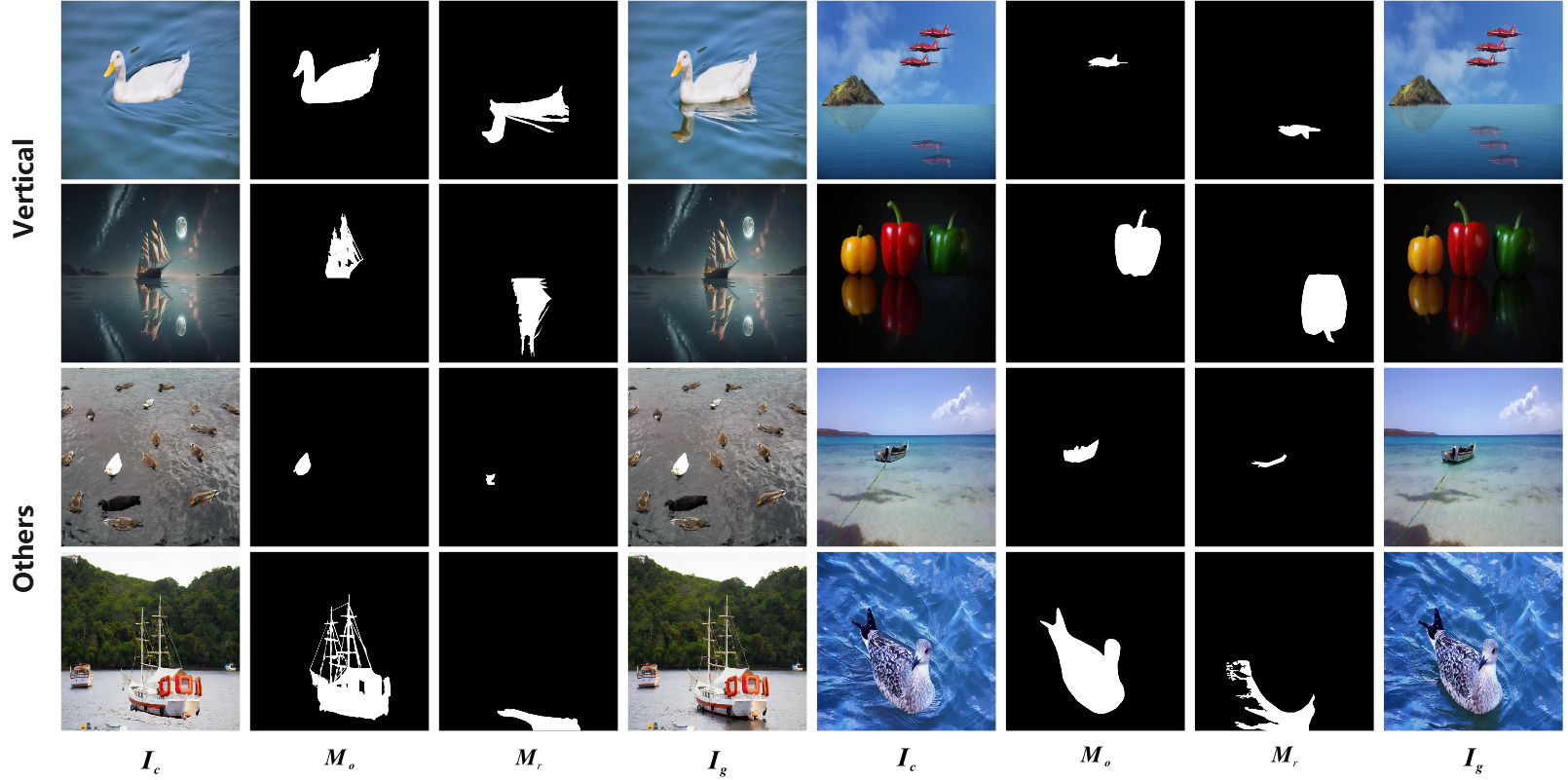}
\caption{Examples from the DEROBA dataset illustrating two types of reflections: ``vertical'' and ``others''. 
The first two rows show samples of the ``vertical'' type, while the last two rows correspond to the ``others'' type. 
For each example, we present a tuple consisting of the composite image $I_{c}$, the foreground object mask $M_{fo}$, the foreground reflection mask $M_{r}$, and the ground-truth target image $I_{g}$.}

\label{fig:type}
\end{figure*}

\section{Discussion on Reflection Type and Bounding Box Prediction}
\label{sec:encoder}
In the main paper, we divide reflections into two types: ``vertical” and ``others”. 
We count these two types of tuples in DEROBA. The ratio of ``vertical" to ``other" types is about 9:1. We visualize some representative examples belonging to both types in Fig.~\ref{fig:type}. The first two rows show the examples belonging to ``vertical” types, whereas the last two rows show the examples from ``others” type.

We train a binary reflection type classifier $E_b$ on the training set of DEROBA. 
We report the classification accuracy on both training set and test set in Table~\ref{tab:cls}. It can be seen that the trained classifier can generally determine the reflection type, laying a solid foundation for generating realistic reflections.

Note that $E_b$ also accounts for predicting the regression coefficients, based on which rotated foreground bounding box can be converted to rotated reflection bounding box. 
We report the accuracy of rotated bounding box regression on both training set and test set in Table ~\ref{tab:bbx}. We adopt KFIoU~\cite{yang2022kfiou} as the evaluation metric. Note that the upper bound of KFIoU is 0.333. Furthermore, we visualize bounding box regression Fig.~\ref{fig:box_supp}, demonstrating that the predicted bounding box can roughly enclose the reflection area. 

\section{More Visual Comparison}
\label{sec:more_result}
\begin{figure*}[t]
\centering
\includegraphics[width=1\linewidth]{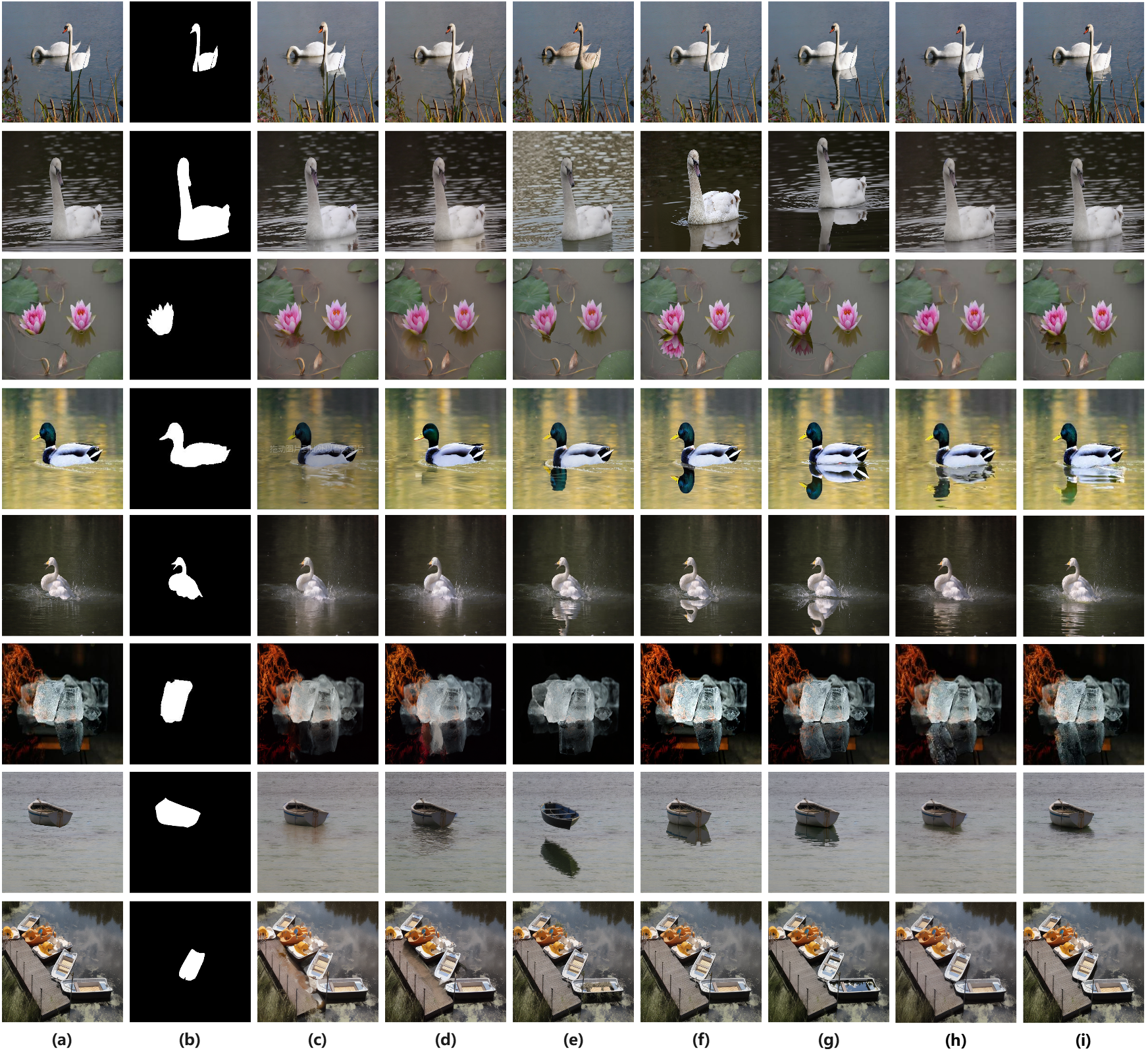}
\caption{Visual comparison of different methods on DEROBA dataset. From left to right are input composite image (a), foreground object mask (b), results of SD-ControlNet \cite{rombach2022high} (c), FLUX-ControlNet \cite{zhang2023adding} (d), ICEdit \cite{zhang2025context} (e), FLUX-Kontext~\cite{batifol2025flux} (f), Qwen-Edit~\cite{wu2025qwen} (g), our method (h), and ground-truth (i).}
\label{fig:vis_deroba_supp}
\end{figure*}

In the main paper, we present visualization results of different methods on DEROBA test set. Here, we provide more visualization results in Fig.~\ref{fig:vis_deroba_supp}. It can be seen that our method performs well across different types of reflective media and different categories of foreground objects. 
It can be seen that ControlNet often generates incomplete and low-quality reflections for the foreground objects (\emph{e.g.}, row 1, 4, 6). ICEdit tends to alter the original information of background (\emph{e.g.}, row 1, 2). FLUX-Kontext and Qwen-Edit exhibit obvious copy-and-paste effect, that is, the generated reflections are simply the copies of vertically flipped foreground objects. The generated reflections are too sharp and thus unrealistic.  

For the reflections belonging to ``others" type (\emph{e.g.}, row 7, 8), the baselines tend to generate vertically flipped reflection or no reflection, while our method can generate realistic reflections close to the ground-truth.

\section{Efficiency Comparison}
\label{sec:time}
In this section, we analyze the inference efficiency including model size and inference time. 
In Table~\ref{tab:param}, we compare our method with two competitive baselines: Qwen-Edit~\cite{wu2025qwen} and FLUX-Kontext~\cite{batifol2025flux}. 
We evaluate each model on all test images on a single NVIDIA A100 and calculate the averaged inference time.
Compared with Qwen-Edit and FLUX-Kontext, our model has significantly fewer model parameters and faster inference speed. The results show that our method is much more efficient, while generating more realistic reflections. 

\begin{table}[t]
    \caption{The comparison of model size and inference time with state-of-the-art open-source image editing models.}
  \centering
  \resizebox{0.7\linewidth}{!}{
  \begin{tabular}{c|cc}
      \toprule[0.8pt]
      Method & Param & Inference Time \\
      \hline
      Qwen-Edit~\cite{wu2025qwen} & 20B & 20s \\
      FLUX-Kontext~\cite{batifol2025flux} & 12B & 13s \\
      \hline
      Ours & 1.3B & 2.8s \\
      \bottomrule
  \end{tabular}
  }
  \vspace{3pt}
  \label{tab:param}
\end{table}

\bibliographystyle{IEEEbib}
\bibliography{icme2026references}